\documentclass[10pt,twocolumn]{article}

\usepackage{arxiv}
\usepackage[utf8]{inputenc}
\usepackage[T1]{fontenc}
\usepackage{microtype}
\usepackage{booktabs}
\usepackage{graphicx}
\graphicspath{{figures/}}
\usepackage{amsmath,amssymb}
\usepackage{xcolor}
\usepackage{enumitem}
\usepackage{array}
\usepackage[numbers,sort&compress]{natbib}
\usepackage{hyperref}
\usepackage{url}

\definecolor{clinicalblue}{HTML}{1F5A7A}
\definecolor{clinicalteal}{HTML}{247A73}
\hypersetup{
  colorlinks=true,
  linkcolor=clinicalblue,
  citecolor=clinicalteal,
  urlcolor=clinicalblue
}

\setlist[itemize]{leftmargin=*,topsep=3pt,itemsep=2pt}
\setlist[enumerate]{leftmargin=*,topsep=3pt,itemsep=2pt}

\title{Temporal Evidence Allocation for Longitudinal Clinical Retrieval:\\
A Position and Technical Report}
\renewcommand{\shorttitle}{Temporal Evidence Allocation for Clinical Retrieval}

\author{
  Soumya Ranjan Sahoo, Divya Bharti, Sanand Sasidharan\\
  GE HealthCare Technology and Innovation Center, India
}

\date{}

\begin{document}
\raggedbottom
% The arXiv style's title box is designed for one-column pages.  In a
% two-column short paper, put the title and abstract in the full-width
% optional argument of \twocolumn so that they cannot collide with the
% running header or the first column.
\twocolumn[
  \maketitle
  \begin{abstract}
Electronic health records are written to support care, not retrieval. Across years of oncology notes, the primary diagnosis is restated in report after report, while the events that shape a patient's course, such as a recurrence or a change in therapy, may be recorded only once. A retriever that ranks passages by similarity to the query keeps returning the repeated material and can miss these transitions, that is, it exhibits temporal redundancy across historical narratives. We argue that longitudinal clinical retrieval should be treated as allocating a limited evidence budget across time, not just ranking by relevance. We separate the task into three objectives: temporal evidence allocation, candidate reconstruction, and clinical validation. In Study A (13 clinical queries over 20 patient records, with two-stage reconstruction), the Temporal-15 configuration reaches a milestone slot recall of 0.538, compared with 0.388 for the Global-50 baseline. However, it also retrieves far more evidence (496 versus 50 indexed units per query), so the gain cannot be credited to allocation alone. When evidence is held fixed, removing prompted precedence rules raises schema-order reversals from 0.010 to 0.080. A five-patient, single-query pilot with fixed budgets finds that wider temporal coverage does not reliably yield more dated milestones. Together, these results suggest that how much a record says about a period is a poor guide to how useful that period is as evidence. Because no expert reference timelines are available, our measurements describe retrieval behavior rather than clinical accuracy. This position and technical report presents the retrieval analysis behind the Onco-CONSTRUE framework and sets out a research agenda for budget-constrained longitudinal clinical retrieval.
  \end{abstract}
  \noindent\textbf{Keywords:} longitudinal clinical retrieval, temporal evidence allocation, oncology timelines, retrieval-augmented generation, no-gold development diagnostics
  \vspace{0.4em}
]

% Self-contained short report: shared sections and the larger manuscript stay unchanged.
\section{Introduction: the missing transition problem}
\label{sec:introduction}

Consider an author-constructed oncology record. Six follow-up notes copy forward a cancer diagnosis. A single earlier pathology report records confirmation, and a later scan suggests progression. With a small evidence allowance, a global similarity search might spend most of its selections on the repeated notes. A time-aware allocation could give the isolated reports a chance to enter the context. If the model then places treatment before confirmation, a prompted precedence check could flag or discourage that order under the declared milestone schema. This example is illustrative, not an observed failure or measured benefit. It shows three separate decisions: where to search, which evidence adds information, and how to order candidate events.

Electronic health records contain the materials for a timeline, but rarely contain the timeline itself. Pathology, radiology, treatment records, laboratory results, and narrative notes describe different parts of a patient's course. Some notes restate old events. Others describe a plan that was never carried out. A useful reconstruction must connect these observations without treating repetition as new evidence.

The missing transition is an IR problem before it is a generation problem. Diagnosis, treatment changes, and recurrence may appear only once among years of routine documentation. Semantic relevance alone does not tell a system how to spread its attention across that history. Longer context does not settle the issue either: models can underuse information in long inputs~\citep{liu2024lost}, and longitudinal clinical tasks still require temporal reasoning~\citep{kruse2025temporal}.

Our position is that longitudinal retrieval is \emph{evidence allocation}. Time, source diversity, repetition, and the question being asked should help determine which evidence enters a finite context. Density describes the observation process, but not necessarily clinical importance. The next question is not whether to retrieve from dense periods, but when evidence from another period adds new clinical information.

This report develops the methodological foundation emerging from Onco-CONSTRUE. It extends the retrieval analysis underlying our accepted IEEE BHI 2026 abstract. That abstract describes a broader architecture, including Clinical Skill Units (CSUs), temporal governance, an independent validation stage, retry logic, endpoint computation, and privacy controls. Here we examine retrieval allocation and prompt-level candidate reconstruction, rather than the complete framework. The contribution is a task formulation, a measurement framework, and lessons from two studies, not a new oncology extraction model.

\textbf{Study A is the main mechanism exploration}: 20 patients, 13 clinical queries, two-stage reconstruction, four conditions, and three runs per condition. It examines retrieval exposure and a paired constraint ablation. \textbf{Study B is a secondary budget-controlled pilot}: five patients, one whole-course query, direct reconstruction, five policies, and three indexed-unit limits. It asks whether spreading the same number of indexed units across more periods also fills more milestone slots. \textbf{Study B is not a matched reanalysis of Study A.} It changes the task and does not test the effectiveness of the 13-query pipeline.

\begin{center}
\setlength{\fboxsep}{7pt}
\fbox{\begin{minipage}{\dimexpr\columnwidth-2\fboxsep-2\fboxrule\relax}
\small\textbf{What this paper claims and does not claim}

\medskip\textbf{Claims.} Temporal evidence allocation should be an explicit IR objective. Retrieval and chronological reconstruction are distinct stages. Prompted precedence rules reduce schema-order reversals in a paired comparison. No-gold diagnostics expose different failure modes. Broader temporal coverage does not guarantee more dated milestones.

\medskip\textbf{Does not claim.} Clinically accurate patient timelines, validated progression or treatment-line endpoints, superiority of a specific retriever at a matched budget, generalization beyond this cohort, or production and regulatory readiness.
\end{minipage}}
\end{center}

\section{Position: longitudinal retrieval is evidence allocation}
\label{sec:position}

The framework separates five questions:
\begin{itemize}
\item \textbf{Temporal evidence allocation:} where retrieval capacity is spent across the patient history.
\item \textbf{Temporal breadth:} how much of the documented history appears in the selected evidence.
\item \textbf{Evidence exposure:} how many indexed units and tokens reach each model stage.
\item \textbf{Reconstruction validity:} whether candidate output follows the declared schema and order.
\item \textbf{No-gold development diagnostics:} whether output behavior and evidence dependence are coherent enough to justify expert review.
\end{itemize}
Allocation chooses where to look, while ranking chooses items within those opportunities. Breadth describes the spread of attention, not event recovery. Exposure should be defined according to the appropriate model stage, as hundreds of retrieved units do not necessarily correspond to hundreds of units included in the final reconstruction prompt. Reconstruction validity concerns the structural integrity of the reconstruction process, whereas confirming whether an event actually occurred remains a clinical matter. No-gold diagnostics help identify failures before expensive annotation, rather than certify an output.

\subsection{Budgeted allocation}
For period $w$, let $N_w$ be its indexed-unit count and $k_w$ the number of units selected. A fixed indexed-unit allowance $K$ imposes
\begin{equation}
0\leq k_w\leq N_w,\qquad \sum_w k_w\leq K.
\label{eq:budget}
\end{equation}
Rank indexed units within each period by query similarity. Assign each period a weight, convert the weights to quotas, cap quotas at available units, and redistribute unused quota. Section~\ref{sec:study-b} specifies the tested rules. A clinically meaningful weight does not necessarily increase in proportion to the number of indexed units. Repeated information may contribute little additional value, whereas a relatively sparse period may capture a critical transition or key change.

\subsection{Three related research conversations}

\paragraph{Longitudinal patient-record retrieval.}
Clinical IR links patient-specific questions to evidence~\citep{roberts2016trec}, and RAG supplies that evidence to a generator~\citep{lewis2020rag}. Within record retrieval must also handle private, heterogeneous, and redundant histories. MedAlign grounds clinician-written instructions in longitudinal EHRs~\citep{fleming2024medalign}. Query-dependent truncation and event and time aware EHR retrieval show why selection rules deserve study~\citep{chouhan2025fixedk,cao2026ehrrag}.

\paragraph{Temporal events and timeline reconstruction.}
Temporal IR distinguishes when a document was written from when its events occurred~\citep{campos2014temporal}. Clinical temporal NLP separates event extraction, date normalization, and relation prediction~\citep{miller2023endtoend}. ChemoTimelines makes oncology reconstruction a concrete evaluation task~\citep{yao2024chemotimelines,yao2025chemotimelines}. Cross-modal work uses structured rows to calibrate narrative events~\citep{kumar2026text}. We hold that upstream representation fixed and examine which evidence reaches reconstruction.

\paragraph{Evaluation without full clinical labels.}
Attribution, abstention, and completeness measure different aspects of generated clinical text~\citep{buchmann2024attribute,xie2024doclens,nagar2025umedsum}. They motivate separate diagnostics rather than a single plausibility score. Downstream-aware evaluation such as eRAG offers a useful IR precedent~\citep{salemi2024erag}, but requires downstream ground truth. No-gold development diagnostics measure inspectable behavior while leaving clinical correctness for later validation.

\section{Oncology reconstruction task}
\label{sec:task}

\subsection{Data and consolidated evidence layer}
The internally held, de-identified source cohort contains 140 pseudonymized oncology patients, 1,235,872 OMOP records, and 14,961 clinical note or report records. Study A uses a prespecified, cost-controlled subset of 20 patients: ten with lung cancer and ten with prostate cancer, selected across follow-up, recurrence status, and documentation density. Its frozen derived index contains 209,052 dated rows, with a median of 9,435 per patient (range 924--32,376). The recorded follow-up field spans 17 - 88 months. This rage does not limit or truncate longer date spans present in the index.

Notes and report findings contribute atomic facts organized into OMOP-style event categories. Structured OMOP rows contribute complementary observations. The consolidated patient index stores event dates, source types, indexed-unit identifiers, text representations, and embeddings. This follows the broader goal of representing narrative NLP outputs alongside structured observations~\citep{keloth2023omop}. ``OMOP-style'' describes the organization of extracted facts, not demonstrated complete CDM conformance. Raw clinical text and patient identifiers are not released.

\paragraph{The retrieval unit:} An \emph{indexed evidence unit} is one retrievable row in the consolidated index. It represents an OMOP-derived structured observation or an atomic fact derived from narrative material. It is not necessarily a complete note or an independent clinical event. A \emph{source record} is the underlying note, report, or structured observation from which indexed content originates. More than one unit can originate from a narrative source.

An \emph{indexed-unit identifier} uniquely identifies a retrievable row. The implementation calls this field \texttt{id} and renders it as \texttt{source\_id} in prompts. Retrieval deduplicates these IDs, not underlying notes or reports. A distinct indexed ID therefore does not establish a distinct source record. We use ``distinct units'' for the cross-query union counts below and are not report deduplicated at the level of the parent clinical source.

\subsection{Why 13 focused clinical queries?}
The main pipeline searches separately for chemotherapy, surgery, immunotherapy, prostate-specific antigen, radiology, biomarkers, clinical findings, pathology, staging, comorbidities, radiation, toxicity, and performance status. These 13 prespecified views serve different evidence needs. Pathology may establish confirmation, treatment records may establish exposure, and scans may describe disease status. Toxicity and performance-status notes can help interpret a treatment change. One whole-course request asks a single ranking to serve all those needs at once.

Focused queries make the search more directed. They give less frequently documented evidence types their own opportunity to enter the context, rather than requiring them to compete with every repeated description of the dominant diagnosis. Each query produces a source linked evidence timeline. Reconstruction combines the 13 views so that conflicts and missing information can be inspected across sources. This is our clinical design rationale. We did not run a controlled comparison proving that 13 queries outperform one.

\subsection{Candidate milestones and generation}
The 13 timelines feed 11 slots - first evidence, confirmation, three treatment-line starts, three treatment-line ends, and three disease-status transitions. The transition slots retain the legacy recurrence/progression field names. They are not adjudicated endpoints, and the study does not distinguish radiographic, biochemical, and clinical progression. The shared fields do not make recurrence and progression interchangeable.

A slot can contain an ISO date or null. It also records status (documented, approximate, inferred, or unknown), confidence, indexed-unit identifiers, supporting evidence, and a reason. The prompt permits abstention instead of forcing a date.

Queries are embedded with a private deployment of \texttt{text-embedding-3-large}. A private enterprise GPT-5 deployment (version 2025-08-07) generates the evidence views and candidate milestones. The constrained prompt supplies ten adjacent, task-specific precedence relations used in this exploratory schema. They operationalize the tested reconstruction format. They are not proposed as universal oncology chronology. These are model instructions, not a deterministic temporal solver.

\begin{figure*}[t]
\centering
\includegraphics[width=\textwidth]{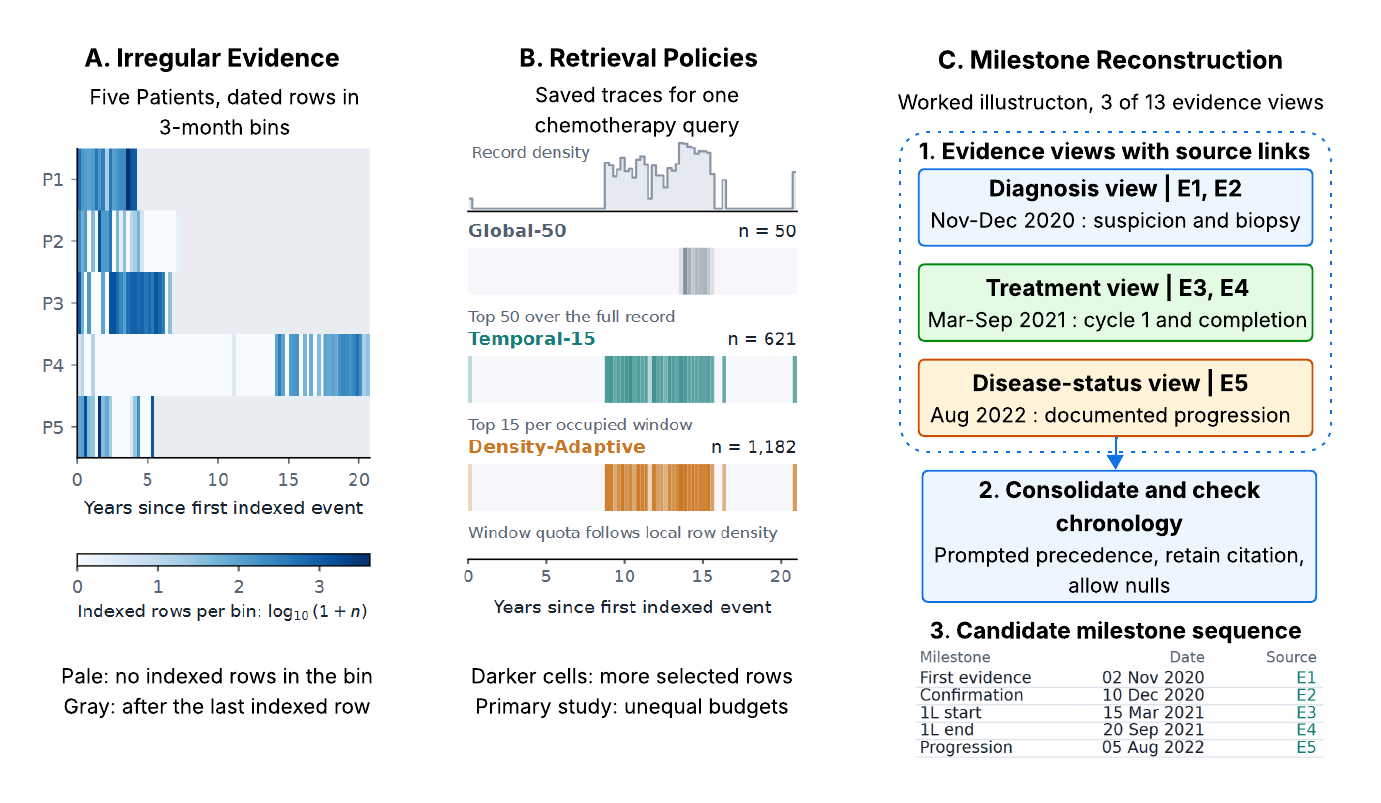}
\caption{\textbf{From irregular records to candidate milestones.} (A) Indexed-unit counts in three-month periods for five de-identified patients, aligned to their first indexed event. Gray cells follow the last indexed unit; color shows $\log_{10}(1+\mathrm{units})$. (B) Saved selections for one chemotherapy query in Study A, and counts differ across policies. (C) An author-constructed example links dated evidence views to prompted chronology checks and candidate milestones. E1--E5 are illustrative source links. 1L is the first systemic treatment line.}
\label{fig:short-overview}
\end{figure*}

\begin{figure*}[t]
\centering
\includegraphics[width=\textwidth]{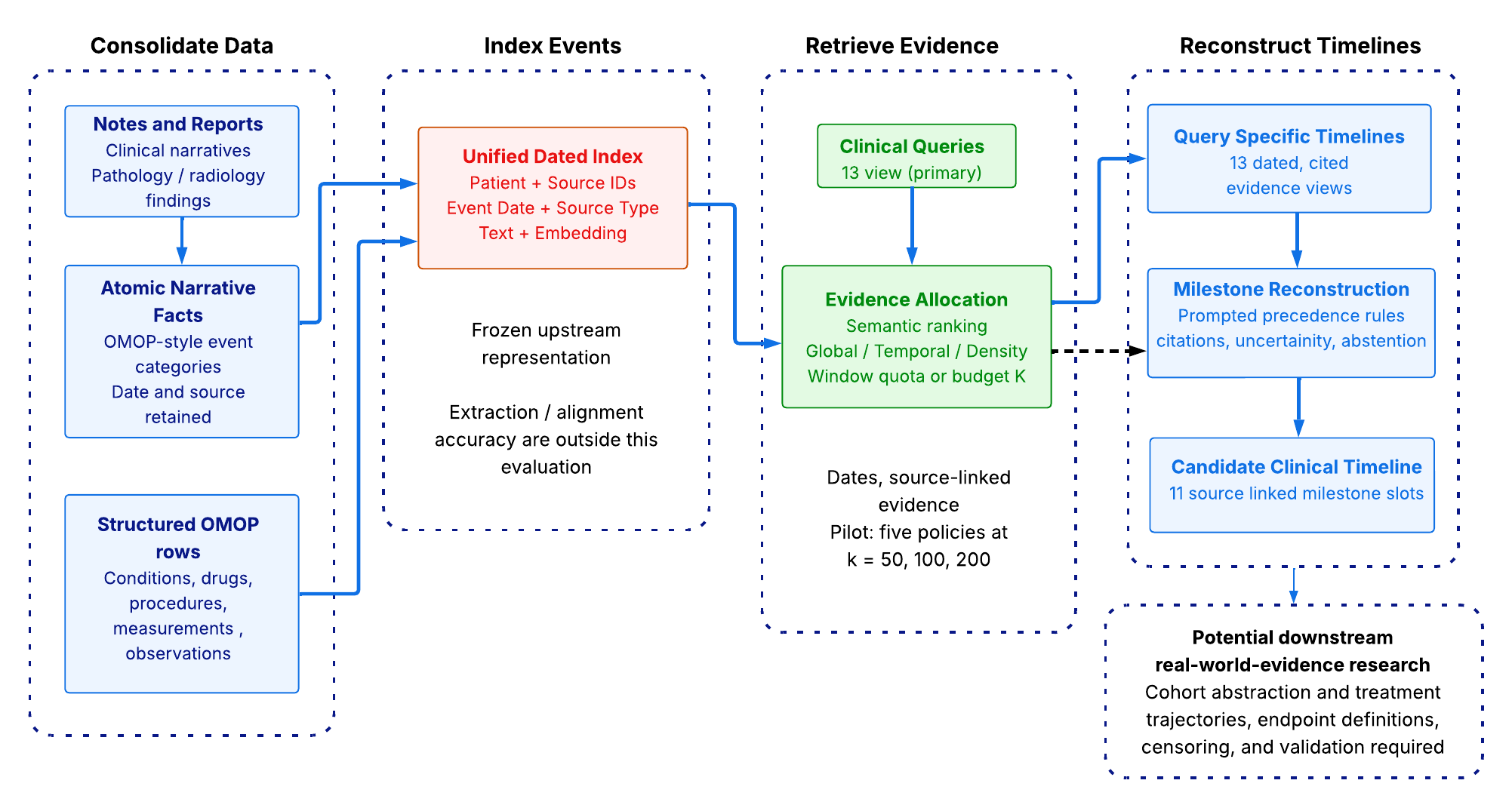}
\caption{\textbf{The data-to-timeline framework.} Narrative facts and structured OMOP observations form a dated, source-linked index. Study A retrieves evidence for 13 clinical queries, generates their evidence views, and reconstructs 11 candidate milestones. The dashed bypass shows Study B's direct reconstruction from one whole-course retrieval request. The downstream box describes intended real-world-evidence research, not an evaluated endpoint pipeline.}
\label{fig:methodology}
\end{figure*}

\subsection{What reaches each model stage}
Each selected unit enters a clinical-query prompt as its indexed ID, event date, and stored text representation. A bracketed header contains \texttt{source\_id=UNIT\_ID} and \texttt{event\_date=YYYY-MM-DD}, followed on the next line by the unit's text. Structured observations and narrative facts share this wrapper; their text lengths can differ.

For windowed retrieval, selections are unioned and deduplicated by indexed-unit ID before the query prompt is built. The means of 496 and 965 units per query count that deduplicated selection. The runner includes every selected unit in its query prompt, without application-side unit thinning at this stage. Inclusion in a submitted prompt does not show that the model used every item.

The query model compresses those units into one generated evidence view per clinical query. Milestone reconstruction receives all 13 views, plus a metadata-only catalog of the union of selected units across queries. Catalog entries contain the indexed ID, event date, and event type, not the original text. Above 2,500 distinct units, the catalog is thinned to 2,500 evenly spaced entries over sorted IDs. The 13 views are not thinned by this rule. The cap applies to all selected units in the catalog, not only to units cited in the views.

\begin{quote}\small
\begin{minipage}{\linewidth}
Selections $\rightarrow$ ID-deduplicated units\\
$\rightarrow$ query prompts $\rightarrow$ 13 evidence views\\
$\rightarrow$ milestone prompt + metadata catalog.
\end{minipage}
\end{quote}
Study A's retrieval exposure describes the query stage, not the final milestone context length. Study B supplies selected units directly to reconstruction.

\subsection{Three metric families}
\textbf{Output behavior} comprises slot fill (the fraction of 11 slots assigned a date) and abstention (null or unknown slots). We retain ``coverage'' for the historical Study A measure, but mean output completeness, not clinical recall.

\textbf{Temporal validity} measures schema-order reversals among pairs with both dates present. The denominator is essential: missing dates can leave a seemingly well-ordered output largely untested. Study A reports the mean output-level reversal rate, excluding outputs with no evaluable pair, rather than a pooled ratio over all pairs.

\textbf{Evidence behavior} measures indexed-unit linkage and sensitivity to evidence removal. In Study A, date-near support means that a dated slot cites at least one indexed unit within 30 days of its predicted date. Leave-one-citation-out stability checks whether another such unit remains when a citation is removed. It measures redundancy in local support, rather than rerunning the generator. Study B removes indexed units and reruns retrieval and reconstruction. Secondary diagnostics appear in Appendix~\ref{sec:secondary}.

\section{Study A: temporal allocation and constraint ablation}
\label{sec:study-a}

\subsection{Design}
The main study holds the 20-patient index, 13 query texts, scorer, model, and output schema fixed across four conditions. Each has three repeated runs per patient.
\begin{description}[leftmargin=0pt,labelindent=0pt,style=nextline]
\item[Global-50.] Select the 50 highest-scoring indexed units across the patient's history for each clinical query.
\item[Global-50/no-constraints.] Reuse Global-50's evidence and 13 generated views, then reconstruct without the ten precedence rules. This is the paired constraint ablation.
\item[Temporal-15.] Use overlapping three-month windows with a one-month stride. Select up to 15 relevant units per non-empty window, deduplicate indexed-unit identifiers, and order the retained evidence by date.
\item[Density-Adaptive.] Use the same windows but increase the quota with indexed-unit count:
\begin{equation}
k_w=\min\left(40,\max\left(10,\left\lceil0.35N_w\right\rceil\right)\right).
\label{eq:adaptive-k}
\end{equation}
Select all available units if fewer than the quota exist. The floor gives sparse windows an allowance and the ceiling limits dense windows. The coefficient and bounds were fixed engineering choices, not fitted optima.
\end{description}

Each condition contains $20\times3=60$ milestone outputs and $20\times13\times3=780$ query-level timeline artifacts. The no-constraint condition reuses Global-50's timeline artifacts rather than generating new views. Primary results are descriptive means over 20 patients and three runs. Because runs are nested within patients, Table~\ref{tab:short-primary} does not treat the 60 outputs as independent observations. The runner checkpointed outputs and saved prompt hashes and retrieval traces. Appendix~\ref{sec:secondary} reports a separate five-patient sensitivity analysis within Study A, and it is not Study B.

\subsection{Main results}
\begin{table}[t]
\centering\scriptsize
\setlength{\tabcolsep}{3pt}
\renewcommand{\arraystretch}{1.12}
\caption{Study A: descriptive means over 20 patients and three runs. Higher values indicate more fill, date-near cited support, and citation-removal stability, lower schema-order reversal rates are preferable. Fill and abstention are complementary in these outputs, not two accuracy measures. Support means a cited unit lies within 30 days of the assigned date, not semantic entailment. Stability means another date-near cited unit remains after one citation is removed.}
\label{tab:short-primary}
\begin{tabular}{lrrrrr}
\toprule
Policy & Fill & Reversals & Support & Abstain & Stability\\
\midrule
Global-50 & .388 & .010 & .959 & .612 & .586\\
No-constraints & .405 & .080 & .981 & .595 & .600\\
Temporal-15 & .538 & .000 & .951 & .462 & .801\\
Density-Adaptive & .553 & .011 & .936 & .447 & .821\\
\bottomrule
\end{tabular}
\end{table}

Temporal-15 fills more slots than Global-50: 0.538 versus 0.388 (Table~\ref{tab:short-primary}). More date-near support also survives single-citation removal: 0.801 versus 0.586. This comes with much greater exposure. Mean indexed units per query rise from 50 to 496, and mean distinct units across the 13 queries rises from 244 to 1,522.

Density-Adaptive increases exposure to 965 indexed units per query and 2,739 distinct units across queries. Yet fill rises only from 0.538 to 0.553, and dropout stability from 0.801 to 0.821. Date-near support falls from 0.951 to 0.936. Repeated-run date agreement also falls (Appendix~\ref{sec:secondary}). Almost doubling the evidence relative to Temporal-15 does not produce comparable diagnostic gains.

The paired ablation provides the cleanest controlled contrast in Study A. Removing precedence instructions while reusing Global-50's evidence and views raises the mean schema-order reversal rate from 0.010 to 0.080. Across all outputs, 163, 164, 260, and 267 pairs are evaluable for Global-50, no-constraints, Temporal-15, and Density-Adaptive, respectively. These are denominator counts, not the denominator of a pooled Table~\ref{tab:short-primary} rate. Retrieval and reconstruction safeguards therefore address different problems. The effect concerns compliance with the declared milestone order.

Across all 20 patients, we first average the three runs within patient and then subtract paired fill values. Temporal-15 minus Global-50 has a median difference of 0.121 (IQR 0.000--0.197): 14 patients improve, five tie, and one declines. Density-Adaptive minus Temporal-15 has a median of 0.000 (IQR 0.000--0.030): seven improve, nine tie, and four decline. These distributions describe heterogeneity rather than a significance test; quartiles use linear interpolation at $(n-1)q$. The aggregate paired summary accompanies the source package.

\section{Study B: evidence budgets}
\label{sec:study-b}

\subsection{Purpose and single-query protocol}
This secondary pilot asks: \emph{with the same number of indexed units, does evidence from more periods lead to more dated milestones?} Five density-stratified patients were previously frozen for an independent-model audit. For each patient, five policies retrieve 50, 100, or 200 distinct indexed units. The model reconstructs the 11 milestones directly, using precedence instructions and allowing null dates.

The exact retrieval query in the frozen configuration is:
\begin{quote}\small
``Retrieve the evidence needed to reconstruct the longitudinal oncology trajectory, including first cancer evidence, pathological confirmation, starts and ends of successive systemic treatment lines, and subsequent progression or recurrence.''
\end{quote}

One whole-course request makes the allowance easy to define: $K$ indexed units for one patient request, rather than 13 allowances followed by overlapping evidence views. This simplifies a budget experiment. It does not preserve the main pipeline's clinical decomposition. Study B changes the queries, uses non-overlapping windows, and bypasses the 13-view stage. Its fill values should not be compared directly with Study A's or used to judge the benefit of focused queries.

\subsection{Five policies in plain language}
\begin{description}[leftmargin=0pt,labelindent=0pt,style=nextline]
\item[Global top-$K$.] Take the indexed units most similar to the whole-course query, regardless of date.
\item[Global MMR.] Prefer relevant units while discouraging similar selections. We use maximal marginal relevance~\citep{goldstein1998mmr} with $\lambda=0.5$, the top $2K$ candidates, and min--max-normalized query scores.
\item[Temporal-equal allocation.] Spread the allowance roughly evenly over three-month periods containing indexed units. Reserve one unit per period when feasible, then distribute the remainder. If there are more periods than units allowed, sample periods across the history.
\item[Density-proportional allocation.] Give more of the allowance to periods containing more units. A period's weight is its indexed-unit count $N_w$.
\item[Relevance-weighted temporal allocation.] Give more selections to periods that contain more units and appear more relevant to the query. Soften density with a square root, then adjust it using the relevance of the period's strongest units. This is a fixed rule, not a model that learns or conducts follow-up searches.
\end{description}

\begin{figure*}[t]
\centering
\includegraphics[width=\textwidth]{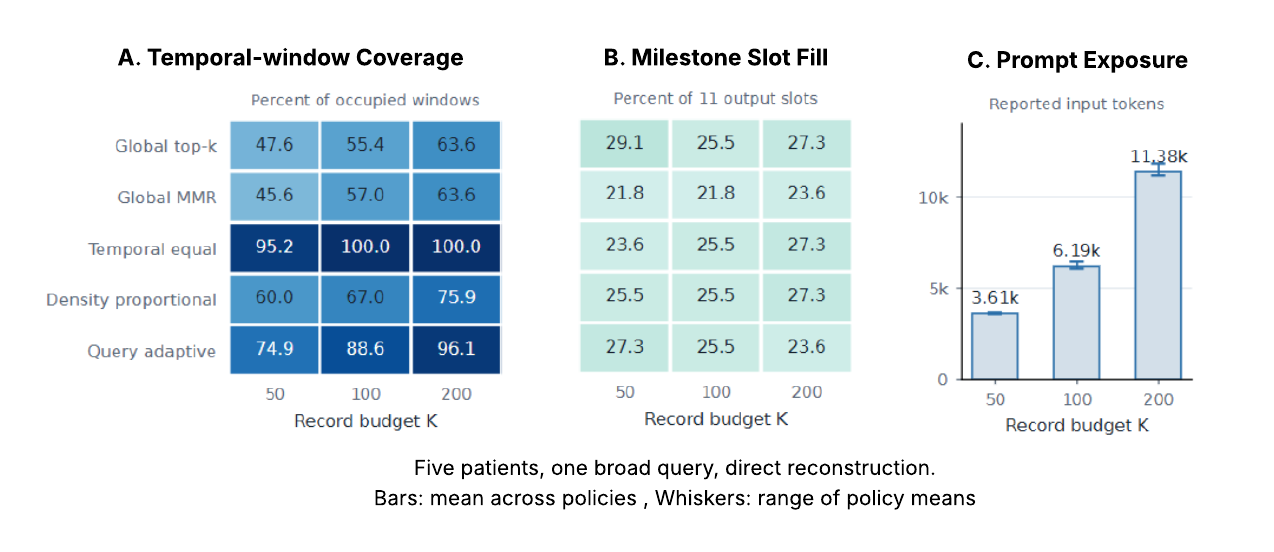}
\caption{\textbf{Study B: selecting evidence under fixed indexed-unit limits in five patients.} Each method selects 50, 100, or 200 indexed units, and the graphic's ``Record budget K'' counts units, not full notes. (A) Percentage of three-month periods containing units from which at least one unit is selected. (B) The percentage of the 11 milestones assigned a date reflects output completeness rather than correctness. (C) Model input length in tokens. Bars represent the average of the five policy means, while whiskers indicate the minimum and maximum policy means rather than statistical uncertainty. ``Query adaptive'' denotes relevance-weighted temporal allocation. All methods use one whole-course query and reconstruct milestones directly. Selecting units from more periods does not consistently produce more dated milestones.}
\label{fig:budget-pilot}
\end{figure*}

The last three policies use non-overlapping three-month periods starting at the first indexed event. They rank units by query similarity within each period. For relevance-weighted allocation the weight is
\begin{equation}
a_w=\sqrt{N_w}\left(1+\frac{r_w}{W}\right),
\label{eq:relevance-weight}
\end{equation}
where $W$ counts periods containing indexed units and $r_w\in\{1,\ldots,W\}$ ranks the average similarity of each period's best three available units. Higher relevance receives a higher rank. Capacity-constrained largest-remainder allocation converts weights to quotas: assign whole-number shares, cap them at available units, and redistribute the remainder. All observed conditions return exactly $K$ units. The saved name is \texttt{query\_adaptive}; the unchanged Figure~\ref{fig:budget-pilot} labels it ``Query adaptive.'' Neither this policy nor density-proportional allocation is Study A's overlapping-window Density-Adaptive method.

\subsection{Measurements and completion}
Temporal breadth is
\begin{equation}
B=\frac{\#\{w:k_w>0\}}{\#\{w:N_w>0\}}.
\label{eq:breadth}
\end{equation}
The denominator excludes periods with no indexed units. If units exist in 20 periods and retrieval selects at least one from 15, temporal breadth is 75\%. Slot fill counts requested milestones receiving dates. API-reported input tokens measure exposure. The baseline comprises $5\times5\times3=75$ patient--policy--budget conditions. Retrieval is rerun under five fixed dropout seeds, each removing approximately 10\% of indexed units: 375 perturbed retrieval sets. Generation is performed for the 75 baseline conditions and once per condition using seed 37, producing 150 generation artifacts in total. All 150 completed successfully.

\subsection{Results: spreading evidence does not reliably fill more slots}
Temporal-equal represents 95.2\%, 100\%, and 100\% of documented periods at allowances of 50, 100, and 200 units. Global top-$K$ represents 47.6\%, 55.4\%, and 63.6\%. Density-proportional represents 60.0\%, 67.0\%, and 75.9\%; relevance-weighted temporal allocation represents 74.9\%, 88.6\%, and 96.1\%. Temporal-equal's breadth partly follows from its design: it explicitly spreads units across periods.

These breadth differences do not yield a consistent ordering of slot fill. Across 15 policy--budget cells, fill ranges from 21.8\% to 29.1\%. Global top-$K$ has the highest observed fill at 50 units, four methods tie at 100, and three tie at 200. Increasing the allowance does not consistently improve fill. Five patients cannot establish equivalence, but the observed results do not support a consistent winner.

Mean input length across policies rises from approximately 3.6k to 6.2k to 11.4k tokens at the three allowances. More input produces no consistent fill increase in this protocol. No schema-order reversal is observed, but only 6--10 precedence pairs can be evaluated per policy budget cell across all five patients. Missing dates leave most relations untested. Secondary perturbation results appear in Appendix~\ref{sec:secondary}.

\section{What the two studies jointly show}
\label{sec:joint}

\subsection{Documentation density is not evidence utility}
Study A gives the central design lesson. A density-based quota nearly doubles exposure relative to a fixed window quota, with modest changes in fill and citation removal stability. Support and repeated run agreement do not improve. Study B reaches a complementary conclusion under a simpler protocol, showing that selecting units from a greater number of periods does not consistently increase slot coverage.

A busy period can contain repeated surveillance, copied-forward diagnoses, billing rows, or multiple representations of one event. A sparse period can hold the report that changes the interpretation of the course. Our data do not identify which mechanism caused each result. They do show why indexed-unit count should not be treated as marginal clinical value.

The next allocation rule should ask whether another unit adds relevant information, covers a missing period, resolves a conflict, or repeats what the model has already seen. Relevance, temporal novelty, redundancy, and task need are candidate signals. They are proposals for the next comparison, not validated components of the present system.

\subsection{Search breadth and ordering solve different problems}
Retrieval creates access to evidence. Reconstruction assigns events and dates. A source can enter the context without being used correctly, and a well-ordered output can omit the decisive transition. The paired constraint ablation shows that reconstruction instructions change schema order even when evidence is fixed. Study B shows that breadth and fill can move separately. Neither property can substitute for the other.

The 13 views preserve clinically directed searches before a model combines the course. The single query pilot intentionally removes that stage for a simpler budget question. A future experiment should compare these query designs while holding the remaining pipeline and total allowance fixed.

\subsection{A compact evaluation checklist}
A no-gold diagnostic protocol should answer five questions:
\begin{enumerate}
\item \textbf{What entered the context?} Report indexed-unit and token exposure, temporal breadth, and query structure.
\item \textbf{What did the model state or leave unknown?} Report slot fill and abstention together.
\item \textbf{What order was tested?} State the schema, evaluable-pair count, and reversals.
\item \textbf{What supports the output?} Distinguish citation membership, date-near support, and semantic support.
\item \textbf{What happens when evidence or generation changes?} Separate indexed-unit removal, evidence ordering, and repeated-run agreement.
\end{enumerate}
The checklist makes results easier to inspect and compare. It also exposes scores that rest on missing dates or weak provenance checks.

\section{Clinical boundary and limitations}
\label{sec:limitations}
The claims boundary in the introduction applies throughout. Candidate timelines may help reviewers locate evidence, but validated progression and treatment-line endpoints require definitions and adjudication. Real-world progression-free survival also requires treatment indexing, progression-or-death rules, death linkage, and censoring~\citep{griffith2019rwprogression,fda2024rwd}. Those endpoints and downstream decisions are outside this evaluation.

The cohort is small, internally held, and limited to lung and prostate cancer. Expert reference timelines are unavailable. A filled slot may contain an incorrect date, and a nearby citation may not be sufficient to verify that the event occurred. Upstream extraction, OMOP alignment, and deduplication are held fixed and unvalidated here. The broader Onco-CONSTRUE architecture is outside the evaluated scope.

Study A's retrieval conditions are not budget matched: their differences combine allocation and volume. Study B controls indexed units within its own protocol, not tokens, and changes queries, window overlap, and reconstruction. It cannot remove Study A's volume confound. Its five patients and one baseline generation per condition limit comparison; it has no power analysis or basis for an equivalence claim.

Both studies use one private model deployment. An open-model audit was incomplete at manuscript freeze and contributes no results. The 2,500 entry milestone metadata catalog cap described in Methods can omit an indexed ID retained in a generated view, complicating provenance in higher-volume conditions. Saved retrieval traces remain complete. Study A's token estimates are calculated by dividing character counts by four, whereas Study B uses API-reported token usage. Exact primary study latency and cost were not consistently saved.

The chronology schema is task specific. Progression can occur before treatment ends, and treatment intervals can overlap. A reversal under the chosen adjacent order is schema noncompliance, not necessarily a clinically impossible sequence. Partial orders and endpoint specific relations need separate evaluation. The pilot's small pair counts and weaker provenance checks are documented with its results and in Appendix~\ref{sec:secondary}, not treated as strong validation signals.

\section{Research agenda}
\label{sec:agenda}

\subsection{Keep the main pipeline fixed when testing budgets}
The next comparison should retain the 13 queries, window geometry, evidence order, and reconstruction prompt while controlling the total distinct units and tokens supplied. It should compare uniform, density-based, relevance-weighted, and redundancy-aware allocation. A separate experiment should compare one whole-course request with focused clinical views at the same total allowance. These tests would isolate questions the present protocols cannot answer.

\subsection{Use controlled cases before full annotation}
Several tests need no complete clinical timelines. Synthetic dated chains can test order and abstention. Removing the only supporting source can test whether the model withdraws a date. Conflicting dates can test whether it reports uncertainty. Duplicate insertion can test whether repetition creates false confidence. Uniform date shifts, with index relations preserved, can test whether reconstructed dates shift consistently. These are proposed tests; success on constructed cases would still need confirmation on clinical records.

MedAlign and ChemoTimelines offer complementary tasks for designing these evaluations~\citep{fleming2024medalign,yao2024chemotimelines,yao2025chemotimelines}. They should inform benchmark design rather than be assumed to provide a matching reference for this multimodal 11-slot trajectory. Expert review remains valuable for later event-level validation. The methodological argument does not depend on completing that annotation first.

\subsection{Make the measurement framework reusable}
The source package includes aggregate figure data, the public pilot audit, and plotting code. Equations~\ref{eq:budget}--\ref{eq:breadth} specify the budget constraint, relevance weighting, and breadth measure, and the pilot query is quoted verbatim. These materials let readers reproduce the graphics and inspect the measurement choices without patient records.

Prompt and schema templates are useful next release items, subject to institutional policy and IP review. This package does not release the clinical index or private prompt corpus. Any release should document query texts, permitted unknowns, precedence relations, source-link rules, and perturbation denominators. The accepted BHI abstract provides the broader Onco-CONSTRUE provenance, whereas this report presents the more focused retrieval argument and the completed studies. They should be cited for their distinct scopes.

\section{Conclusion}
\label{sec:conclusion}
Longitudinal clinical retrieval must decide how to spend attention across a patient history. Study A shows more milestone fill at much greater temporal exposure and isolates the effect of prompted precedence rules. Its density-based expansion adds much more material for modest diagnostic gains. Study B shows that broader coverage alone does not consistently fill more milestones under a simplified indexed unit budget protocol.

Documentation density is not evidence utility. A useful next system must decide when an evidence unit adds new clinical information, not merely when another unit is available. Temporal evidence allocation, separate reconstruction checks, and no-gold development diagnostics provide a practical foundation for that work.

\appendix
\section{Secondary diagnostics and protocol details}
\label{sec:secondary}

\subsection{Study A: repeatability, anchors, and sensitivity}
Run repeatability is the fraction of milestone dates present in paired runs that agree within 30 days. It is 0.880 for Global-50, 0.924 for no-constraints, 0.881 for Temporal-15, and 0.858 for Density-Adaptive. Mean pairwise retrieval-set Jaccard overlap is 1.000, 0.9999, and 0.99999 for the three retrieval policies. With index and embeddings frozen, retrieval is effectively deterministic; most residual date variation occurs during generation.

Weak-anchor agreement asks whether a cited unit's event type belongs to a milestone-specific compatible set. It is 0.816, 0.817, 0.768, and 0.770 for Global-50, no-constraints, Temporal-15, and Density-Adaptive. Event type is only a weak anchor: compatibility does not establish the event's meaning or date.

A prespecified five-patient sensitivity sample averages three runs per patient, then uses 20,000 paired patient-bootstrap resamples. It is separate from Study B, although the samples are not required to be disjoint. Coverage is 0.376, 0.485, and 0.497 for Global-50, Temporal-15, and Density-Adaptive. Temporal-15 minus Global-50 is $+0.109$, with a 95\% bootstrap interval $[+0.012,+0.206]$. Density-Adaptive minus Temporal-15 is $+0.012$ $[0.000,+0.024]$. Corresponding dropout differences are $+0.150$ $[+0.035,+0.271]$ and $+0.062$ $[-0.008,+0.170]$. With five patient clusters these are sensitivity summaries, not population-level precision estimates.

\begin{table}[h]
\centering\scriptsize
\setlength{\tabcolsep}{4pt}
\caption{Study A's five-patient sensitivity sample, not Study B. Tokens are character-based input proxies.}
\label{tab:short-volume}
\begin{tabular}{lrrr}
\toprule
Policy & Units/query & Distinct & Tokens\\
\midrule
Global-50 & 50 & 200 & 36k\\
Temporal-15 & 344 & 1,006 & 164k\\
Density-Adaptive & 704 & 1,883 & 313k\\
\bottomrule
\end{tabular}
\end{table}

Across Study A's 1,665 non-empty overlapping windows, median $N_w$ is 107 (IQR 35-424; range 1-8,907). The density quota operates at its minimum in 21.9\% of windows, proportionally in 29.4\%, and at its maximum in 48.7\%. These describe the frozen operating point, not an optimal quota.

\subsection{Study B: perturbation and weaker provenance checks}
Corpus-dropout seeds are 11, 23, 37, 53, and 71, with generation performed only for seed 37. Retrieval-set Jaccard overlap ranges from 0.786 to 0.818 across policy budget means. Baseline evidence follows retrieval order, whereas perturbed evidence follows indexed-unit ID order. Output changes combine removal, reretrieval, and ordering effects.

Exact date agreement after dropout is 81.8--96.4\%, but includes 37--43 double-null agreements among 55 slots per cell. Agreement that both outputs omit a date does not show that a reconstructed event survived. The aggregate audit records both double-null and non-null denominators, and policies are not ranked according to raw agreement.

Citation membership checks whether at least one cited identifier belongs to selected units. Cell means are 90-100\%. The evaluator assigns 1 to an all null output, so means include vacuous cases. Membership is weaker than Study A's date-near check and does not test semantic support. Cell mean API call durations range from 37.3 to 46.8 seconds, and these measurements reflect service observations rather than an efficiency comparison.

\begingroup
\small
\bibliographystyle{unsrtnat}
\bibliography{main}
\endgroup

\end{document}